\documentclass[11pt]{article}

\usepackage[margin=1in]{geometry}

\usepackage{xspace}
\makeatletter
\DeclareRobustCommand\onedot{\futurelet\@let@token\@onedot}
\def\@onedot{\ifx\@let@token.\else.\null\fi\xspace}

\makeatother

\usepackage[accsupp]{axessibility}  

\usepackage[numbers]{natbib}
\usepackage{microtype}
\usepackage{graphicx}
\usepackage{subcaption}
\usepackage{booktabs} 
\usepackage[table,xcdraw]{xcolor} 
\usepackage{multirow}
\usepackage{tabularx}
\usepackage{wrapfig}
\usepackage{amsmath,amssymb}

\usepackage{tcolorbox}
\tcbuselibrary{listings,skins,breakable}

\definecolor{promptbg}{HTML}{F7F7F8}
\definecolor{promptframe}{HTML}{D0D0D0}
\definecolor{systembg}{HTML}{EEF2FF}
\definecolor{systemframe}{HTML}{8B9FD4}
\definecolor{userbg}{HTML}{F0FAF0}
\definecolor{userframe}{HTML}{7CB88B}

\newtcolorbox{promptbox}[1][]{%
    enhanced,
    breakable,
    colback=promptbg,
    colframe=promptframe,
    fonttitle=\bfseries\small,
    left=6pt, right=6pt, top=4pt, bottom=4pt,
    boxrule=0.5pt,
    #1
}

\newtcolorbox{systempromptbox}[1][]{%
    enhanced,
    breakable,
    colback=systembg,
    colframe=systemframe,
    fonttitle=\bfseries\small,
    title=System Prompt,
    left=6pt, right=6pt, top=4pt, bottom=4pt,
    boxrule=0.5pt,
    #1
}

\newtcolorbox{userpromptbox}[1][]{%
    enhanced,
    breakable,
    colback=userbg,
    colframe=userframe,
    fonttitle=\bfseries\small,
    title=User Prompt,
    left=6pt, right=6pt, top=4pt, bottom=4pt,
    boxrule=0.5pt,
    #1
}

\definecolor{mycommentcolor}{RGB}{255, 127, 0}

\usepackage[breaklinks,colorlinks,citecolor=blue,linkcolor=blue,urlcolor=blue]{hyperref}

\usepackage[capitalize,noabbrev]{cleveref}

\usepackage{orcidlink}

\title{Discovering Failure Modes in Vision--Language Models using RL}

\author{
    Kanishk Jain$^{1,2}$ \qquad
    Qian Yang$^{1,2,\ast}$ \qquad
    Shravan Nayak$^{1,2,\ast}$ \\[6pt]
    Parisa Kordjamshidi$^{3}$ \qquad
    Nishanth Anand$^{1,4}$ \qquad
    Aishwarya Agrawal$^{1,2,5}$ \\[10pt]
    \small $^{1}$Mila -- Qu\'{e}bec AI Institute \quad
    $^{2}$Universit\'{e} de Montr\'{e}al \quad
    $^{3}$Michigan State University \\[2pt]
    \small $^{4}$McGill University \quad
    $^{5}$Canada CIFAR AI Chair
}

\date{}

\begin{document}

\maketitle

\let\thefootnote\relax\footnotetext{$^{\ast}$Equal contribution. Correspondence to \texttt{kanishk.jain@mila.quebec}.}

\begin{abstract}
Vision-language Models (VLMs), despite achieving strong performance on multimodal benchmarks, often misinterpret straightforward visual concepts that humans identify effortlessly, such as counting, spatial reasoning, and viewpoint understanding. Previous studies~\citep{tong2024eyes, fu2024blink} manually identified these weaknesses and found that they often stem from deficits in specific skills. However, such manual efforts are costly, unscalable, and subject to human bias, which often overlooks subtle details in favour of salient objects, resulting in an incomplete understanding of a model's vulnerabilities. To address these limitations, we propose a Reinforcement Learning (RL)-based framework to automatically discover the failure modes or blind spots of any ``candidate VLM'' on a given data distribution without human intervention. Our framework trains a questioner agent that adaptively generates queries based on the candidate VLM’s responses to elicit incorrect answers. Our approach increases question complexity by focusing on fine-grained visual details and distinct skill compositions as training progresses, consequently identifying novel failure modes in which VLMs struggle. We demonstrate the broad applicability of our framework by showcasing its generalizability across various model combinations.




\end{abstract}
\section{Introduction}
\label{sec:intro}

Large-scale web training has enabled Vision-Language Models (VLMs) to achieve impressive performance on multimodal benchmarks~\cite{li2024seed, fu2023mme} spanning diverse tasks such as scene understanding, text recognition, and spatial relation reasoning. This progress has driven their increasing adoption in egocentric perception~\cite{grauman2022ego4d, mangalam2023egoschema} and embodied deployment~\cite{zitkovich2023rt, kim24openvla}, where models must reason over first-person video streams and act in open-ended physical environments. Despite such strong performance, benchmark success has proven to be a poor proxy for reliability; these models still fail on individual instances within the very capabilities these benchmarks claim to evaluate, such as low-level visual perception~\cite{rahmanzadehgervi2024vision} and visual attribute and viewpoint understanding~\cite{tong2024eyes}. Moreover, these benchmarks only cover a subset of visual capabilities, leaving entire categories of potential failures untested. This suggests that current benchmarks are designed to measure how well a model performs on known tasks, rather than systematically probing what these models do not know. This raises two linked questions: what failure modes of these models remain undiscovered, and once surfaced, how can they be leveraged to improve the models themselves?

Identifying failure modes of VLMs is an active area of research,
where past research has successfully identified a subset of failure modes. For instance, one study found that VLMs struggle with tasks involving orientation, viewpoint, and perspective~\cite{tong2024eyes}, and another identified challenges in depth perception, reflectance, and multi-view reasoning~\cite{fu2024blink}. Despite their value, these studies are limited for two reasons. First, they manually design benchmarks for failure mode identification, which is time-consuming and labour-intensive. Second, humans play a major role in discovering such failure modes, which limits the coverage of discovered failure modes due to their cognitive biases~\cite{clark2013whatever}. To overcome these challenges, recent work leverages LLMs as questioners to automatically generate questions and datasets for probing VLMs~\cite{li2024seed}. These methods offer a scalable solution, as LLMs can generate diverse, high-quality queries that are not limited by fatigue or cognitive biases, unlike humans. However, they treat the questioner as a static component. These systems employ ``open-loop'' generation, in which the questioner does not adapt to the candidate VLM's responses. Moreover, the discovered failure modes are generic and do not capture the nuances of a specific model's failure modes.

\begin{figure}[h]
    \centering
    \includegraphics[width=\textwidth]{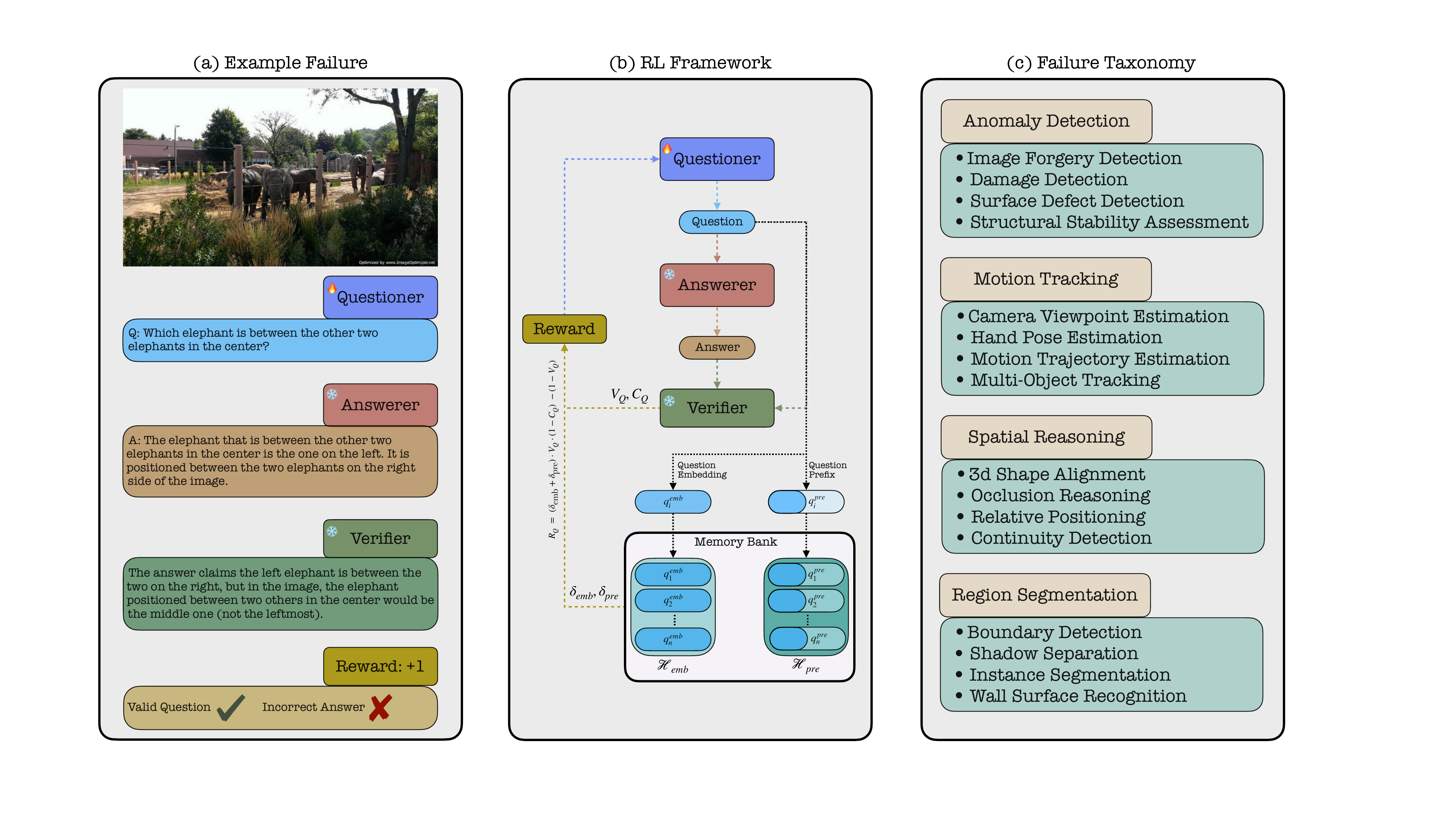}
    \caption{Overview of our framework for automatically discovering failure modes in VLMs. (a) An example failure case, where the questioner generates a valid question which is answered incorrectly by the Answerer (target VLM), (b) In our RL-based framework, the Questioner is trained via reinforcement learning to generate questions that elicit incorrect answers from the Answerer, using a non-stationary reward, (c) A sample of the failure taxonomy automatically constructed from discovered failures.}
    \label{fig:teaser}
\end{figure}

In this work, we introduce a fully automated, RL-based framework in which a VLM-based questioner agent actively learns to identify weaknesses of any target VLM on any image dataset, guided by a reward signal from a verifier VLM. Unlike prior open-loop approaches, the questioner adaptively probes the target VLM throughout training, exploring diverse failure modes while generating increasingly complex questions that probe fine-grained understanding by targeting non-salient objects and capturing model-specific vulnerabilities. To systematically characterise the discovered failures, we introduce a pipeline for creating a failure taxonomy that categorises questions by the cognitive skills required to answer them. The full pipeline is illustrated in Figure \ref{fig:teaser}.

To summarise, we make the following contributions:
\begin{enumerate}
    \item We formalise the task of automated failure discovery by proposing strong baselines and new evaluation metrics that capture the potency and diversity of discovered failures, validated against human judgment.
    \item We propose a novel RL-based framework to automatically discover failure modes of any target VLM on a given dataset. Unlike static benchmarks, our questioner agent adaptively identifies the target VLM's failure modes.
    \item To guide the questioner, we design a non-stationary reward function that encourages generated questions to be both challenging for the target VLM and diverse, preventing the questioner from collapsing onto a narrow set of known weaknesses.
    \item We introduce a pipeline for failure taxonomy construction that categorizes generated questions by the cognitive skills required to answer them, providing structured insight into a model's vulnerabilities.
\end{enumerate}

\section{Related Work}
\label{sec:related_work}

\paragraph{\bf VLM Benchmarking and Failure Probing:} Vision models were traditionally evaluated using aggregate benchmarks~\citep{antol2015vqa}, but more recently, task-specific benchmarks are used to get a fine-grained performance portfolio. For instance, benchmarks such as MME~\citep{fu2023mme} and SEED-Bench~\citep{li2024seed} explicitly separate perception (e.g., OCR, colour, object existence) from cognition (e.g., commonsense reasoning, visual reasoning). Likewise, MMBench~\citep{liu2024mmbench} and MM-Vet~\citep{yu2024mm} further refine these benchmarks to provide comprehensive performance reports in hierarchical and compositional structures, respectively. Similarly, in the failure-mode discovery literature, POPE~\citep{Li-hallucination-2023} evaluates object hallucination via adversarial negative sampling, and HallusionBench~\citep{Guan_2024_CVPR} tests for inconsistencies in visual reasoning. BLINK~\citep{fu2024blink} and MMVP~\citep{tong2024eyes} further isolate fine-grained perceptual gaps, including depth estimation, orientation, and viewpoint dependence in their benchmarks. However, constructing these benchmarks requires substantial human effort for data curation and annotation, thereby limiting the scalability of their approach. Moreover, these benchmarks do not tailor failure modes to a given VLM. In contrast, our framework automatically generates the dataset, eliminating the need for manual intervention.

\paragraph{\bf Automated Probing and Adversarial Policies}: To overcome the costs associated with manual curation, recent works have explored automated probing strategies. In the adversarial robustness literature, such as AttackVLM~\citep{zhao2023evaluating}, pixel-level perturbations are used to break the vision model. However, these methods use noise to expose a model rather than systematically addressing its semantic understanding deficits. More recent works use prompt-based techniques to identify semantic weaknesses~\citep{huang2024conme}, but these approaches are ``open-loop'', as the questioner’s parameters are not updated based on the feedback. In contrast, in our approach, the questioner is trained to adapt its strategy to the target model's specific failure patterns using RL.

Our work parallels adversarial policy research that leverages RL~\citep{gleave2019adversarial, hong2024curiositydriven}, which exploits fixed opponents or red-team LLMs, and EvoVLM~\citep{he2025visplay}, which uses RL to automatically improve VLM's performance. However, our framework focuses on discovering failure modes. In our approach, we leverage a non-stationary reward signal to train the questioner to generate diverse, fine-grained questions that eventually uncover novel failure modes in the target VLMs.

\section{Methodology}
\label{sec:methodology}

\subsection{Problem Formulation}
\label{sec:formulation}

\begin{figure}[h]
    \centering
    \includegraphics[width=0.95\textwidth]{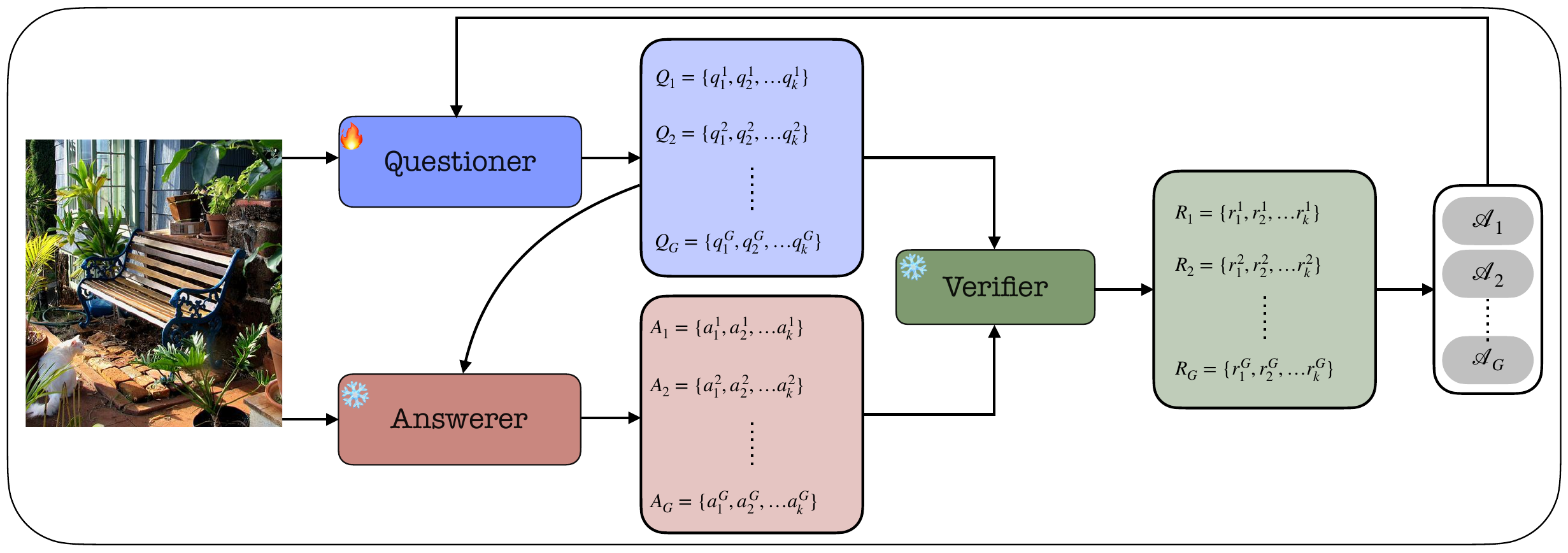}
    \caption{Our approach consists of three models: a \textbf{Questioner} that generates questions, an \textbf{Answerer} that generates answers, and a \textbf{Verifier} that provides the reward signal for training.}
    \label{fig:rl_pipeline}
\end{figure}

We formulate automated failure-mode discovery as a reinforcement learning (RL) problem, motivated by its success on adjacent problems~\cite{shen2025vlm, fan2025grit} and its ability to discover novel concepts beyond the training distribution. Our overall approach, illustrated in Figure~\ref{fig:rl_pipeline}, models failure discovery as a multi-agent interaction among three components: a \emph{Questioner}, an \emph{Answerer}, and a \emph{Verifier}.

\noindent \textbf{Questioner.} The questioner is a VLM-based RL agent that serves as the core of our framework. Given an image $\mathcal{I}$, it generates a set of $k$ questions, $\mathcal{Q} = \{q_1, \dots, q_k\}$, designed to probe the weaknesses of the target VLM. The questioner is the only component whose parameters are updated during training; it learns to adaptively generate questions that elicit incorrect responses from the candidate VLM.

\noindent \textbf{Answerer.} The answerer is the target VLM whose failure modes are being discovered. Given an image $\mathcal{I}$, it generates responses to the questioner's queries. The parameters of the answerer remain frozen throughout training; it serves as the environment against which the questioner is optimized.

\noindent \textbf{Verifier.} The verifier is a reasoning VLM that evaluates the interaction between the questioner and the answerer and provides the reward signal. Given the image $\mathcal{I}$, the generated questions $\mathcal{Q}$, and the answerer's responses $\mathcal{A}$, the verifier assesses two aspects: the \textit{validity} of each question and the \textit{correctness} of each answer.

\begin{figure}[h]
    \centering
    \includegraphics[width=0.85\textwidth]{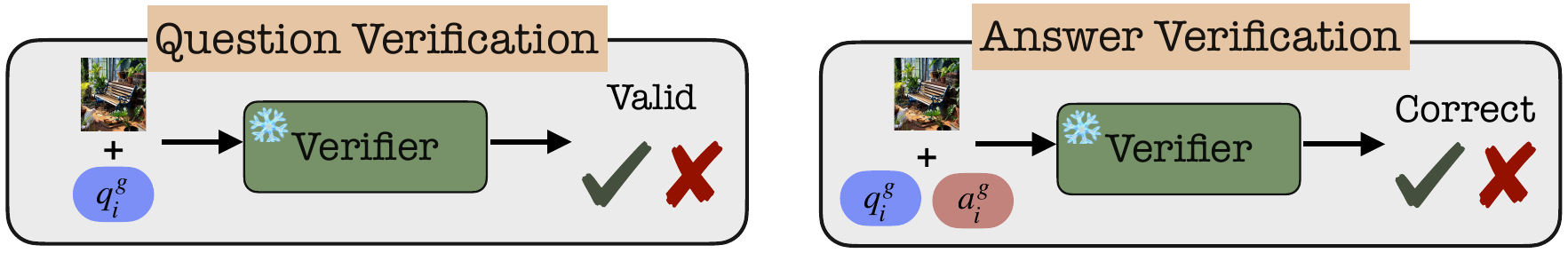}
    \caption{The verification pipeline for question and answer verification.}
    \label{fig:verification}
\end{figure}

To evaluate question \textit{validity}, the verifier assesses each generated question $q_i \in \mathcal{Q}$, conditioned on the image $\mathcal{I}$, across four criteria: (1)~\textit{grammatical correctness}: the question is well-formed; (2)~\textit{atomicity}: the question is standalone and non-compositional; (3)~\textit{visual grounding}: answering the question requires information present in the image; and (4)~\textit{objectivity}: the question admits an unambiguous answer. The validity component $V_Q \in \{0, 1\}$ is set to $1$ only if all four criteria are satisfied; otherwise, $V_Q = 0$. The verifier computes the \textit{correctness} component $C_Q \in \{0, 1\}$ by evaluating the target VLM's response against the generated question and the image. For each component, the verifier first generates chain-of-thought reasoning tokens to internalize the image context, then produces a final judgment along with a justification. The verification process is illustrated in Figure \ref{fig:verification}.

\subsection{Training the Questioner}
\label{sec:training}

\subsubsection{Question Generation}
\label{subsec:ques_gen}

The questioner generates all $k$ questions in a single auto-regressive sequence. Each question $q_i$ is enclosed within distinct tags (e.g., \texttt{<response\_i>}$q_i$\texttt{</response\_i>}) to facilitate downstream processing. Importantly, the questions are generated sequentially within a single pass: each question $q_i$ is conditioned on the previously generated questions $\{q_1, \dots, q_{i-1}\}$, rather than sampled independently. This design prevents redundancy and encourages the questioner to probe distinct visual concepts within the same image, making the failure-discovery process more efficient.

\subsubsection{Optimization}
\label{subsec:optimization}

We train the questioner using the Group Relative Policy Optimization (GRPO) algorithm~\cite{shao2024deepseekmath}. At each update step, GRPO optimizes:
\begin{align}
\mathcal{J}(\pi) = \mathbb{E}_{y_{1:G} \sim \pi_{\text{old}}} \bigg[&\frac{1}{G} \sum_{g=1}^G \frac{1}{|y_g|} \sum_t \bigg(
\texttt{min} \Big( r_t \hat{A}_g, \, \texttt{clip}(r_t, 1-\epsilon, 1+\epsilon) \hat{A}_g \Big) \nonumber \\
&- \beta \mathrm{KL}[\pi || \pi_{\text{ref}}] \bigg) \bigg]
\end{align}
where $\pi$ is the current policy of the questioner, $\pi_{\text{old}}$ is the policy at the end of the previous update, $\pi_{\text{ref}}$ is the reference policy, and $r_t = \tfrac{\pi_{\text{old}}(y_{g,t} | y_{g, <t})}{\pi(y_{g,t} | y_{g, <t})}$ is the probability ratio that constrains the new policy to remain close to the previous one. The clipping parameter $\epsilon$ bounds the ratio, $\hat{A}_g = \tfrac{R_g - \texttt{mean}(R_{1:G})}{\texttt{std}(R_{1:G})}$ is the group-normalized advantage for response $g$, and $\beta$ controls the strength of the KL divergence penalty. We define the reward $R_Q$ below.

\subsubsection{Reward Function}
\label{subsec:reward}

We design a reward signal consisting of multiple sub-components to train the questioner agent:
\begin{equation}
\label{eq:reward}
R_Q \;=\; \Delta_Q \cdot V_Q \cdot (1 - C_Q)\;-P_Q,
\end{equation}
where $R_Q$ is the overall reward for generating question $Q$. The individual components are:
\begin{enumerate}
    \item $V_Q \in \{0, 1\}$, the \textit{validity} component, which ensures the generated question is semantically and grammatically well-formed, and answerable given the image;
    \item $C_Q \in \{0, 1\}$, the \textit{correctness} component, which ensures the questioner is rewarded only when the candidate VLM answers the generated question incorrectly;
    \item $\Delta_Q \in [0, 1]$, the \textit{diversity} component, which scales the positive reinforcement to encourage the generation of diverse question types;
    \item $P_Q \in [0, 1]$, the \textit{penalty} component, which provides negative reinforcement to accelerate training. We set $P_Q = (1 - V_Q)$, thereby penalizing invalid questions.
\end{enumerate}

Intuitively, the reward function provides positive reinforcement when the generated question is valid ($V_Q = 1$), meets the diversity requirement ($\Delta_Q > 0$), and is incorrectly answered by the candidate VLM ($C_Q = 0$). Otherwise, the agent receives no positive reward. Additionally, invalid questions incur a penalty through $P_Q$, which accelerates the learning of well-formed question generation.

We further decompose the diversity component $\Delta_Q$ into semantic and lexical sub-components.

\noindent \textbf{Semantic Diversity.} This term penalizes the generation of questions that are semantically similar to those previously answered incorrectly by the target VLM. For each image $\mathcal{I}$, we maintain an image-level memory bank $\mathcal{H}_{\mathcal{I}}$ of incorrectly answered questions. The semantic diversity score is computed using the cosine similarity between the embedding of the current question $\mathbf{e}(q_i)$ and its nearest neighbour in $\mathcal{H}_{\mathcal{I}}$:
\begin{align}
    \delta_{\text{emb}} = 1 - \max_{q' \in \mathcal{H}_{\mathcal{I}}} \texttt{cos}(\mathbf{e}(q_i), \mathbf{e}(q')),
\end{align}
where embeddings are extracted using the sentence embedding model \textit{all-MiniLM-L6-v2}~\cite{wang2020minilm}. The image-level memory bank encourages localized semantic variety, preventing the questioner from repeatedly probing the same concept within a given image.

\noindent \textbf{Lexical Diversity.} To encourage varied question types (e.g., ``How many...'' vs.\ ``What is...''), we compute the inverse frequency of a question's prefix (the first $L$ tokens) within a global, dataset-level memory bank $\mathcal{H}$:
\begin{align}
    \delta_{\text{ifreq}} = \frac{1}{1 + \texttt{Count}(\texttt{prefix}(q_i), \mathcal{H})},
\end{align}
where $\texttt{prefix}(q_i)$ returns the first $L$ tokens of $q_i$, and $\texttt{Count}(\texttt{prefix}(q_i), \mathcal{H})$ counts the occurrences of this prefix in $\mathcal{H}$. This discourages the questioner from converging on a narrow set of question templates.

The overall reward signal combining both diversity terms is:
\begin{align}
\label{eq:final_reward}
R_Q \;=\; \lambda_{\text{scale}} \cdot (\delta_{\text{emb}} + \delta_{\text{ifreq}}) \cdot V_Q \cdot (1 - C_Q)\; - \lambda_{\text{penalty}} \cdot (1 - V_Q),
\end{align}
where $\lambda_{\text{scale}}$ and $\lambda_{\text{penalty}}$ are scaling factors for the failure discovery and invalid question penalty components, respectively.

\subsection{Categorizing Failure Modes}
\label{sec:fail_mode}

To understand the failure modes of the target VLM, we map the generated questions to the skills and meta-skills required to answer them using a novel clustering pipeline. Our approach leverages LLMs at various stages, and we provide all the details, including the prompts used, in the Appendix. As illustrated in Figure~\ref{fig:failure_taxonomy}, our approach consists of four stages: identification of primitives, topic modeling, extraction of skills, and identification of meta-skills. We briefly describe each step below.

\noindent \textbf{Identification of primitives:} In this step, we identify the primitive skills required to answer a particular question for each method separately. To obtain this, we first select valid questions among all the generated questions. Then, we pass these questions to a VLM along with the corresponding image, prompting it to identify the set of primitive skills required to answer them.

\begin{figure}
  \centering
  \includegraphics[width=0.85\textwidth]{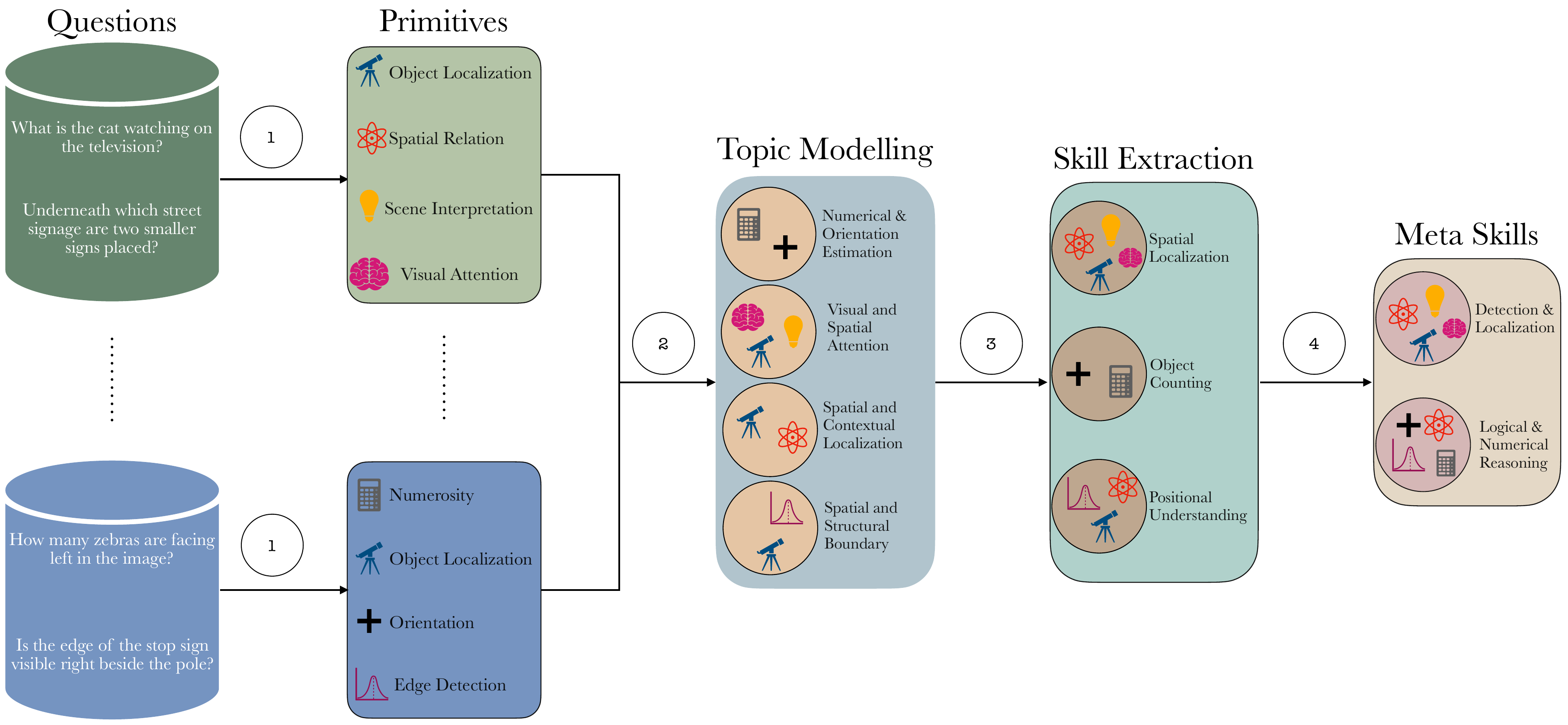}
  \caption{The failure taxonomy pipeline consists of four stages: (1) identification of primitives, (2) topic modelling, (3) skill extraction, and (4) meta-skill identification.}
  \label{fig:failure_taxonomy}
\end{figure}

\noindent \textbf{Topic modelling:} Here, we use topic modelling to map thousands of primitive skills identified across all the questions into a handful of meaningful categories. Unlike performing clustering independently per method, we pool the primitive skills from \textit{all} methods into a single set. We first deduplicate near-identical skills using embedding cosine similarity, then cluster the combined pool using BERTopic~\cite{grootendorst2022bertopic}. For each cluster, we use an LLM to generate a descriptive label, yielding the topic name. This pooled approach produces a shared taxonomy across all methods, enabling direct and fair comparison of skill distributions without the need for post-hoc cross-method standardization.

\noindent \textbf{Skill extraction:} In this step, we refine the topic labels produced by the clustering step. We normalize the LLM-generated labels for consistency and deduplicate near-identical topic labels using embedding similarity. The resulting cluster assignments are then mapped back to each method's questions, so that every method's questions are categorized under the same unified set of skills.

\noindent \textbf{Meta-skill identification:} In this final step, we leverage an LLM to group skills into meta-skills in order for us to have two hierarchical levels of taxonomy for studying failure modes.

\section{Experimental Results}
\label{sec:experiments}

\subsection{Experimental Setup}
\label{sec:setup}

\noindent \textbf{Dataset.} We randomly sample $1000$ images from the training set of the \textsc{COCO}~\cite{lin2014microsoft} dataset to probe the target VLM's failure modes. Our framework is agnostic to the choice of dataset; the failure modes discovered are tailored to the specific target VLM and dataset combination.

\noindent \textbf{Models.} We experiment with a range of target VLMs (answerers) spanning 3B to 8B parameters across two architecture families: \textit{Qwen2.5-VL} and \textit{Qwen3-VL}. For the questioner, we use \textit{Qwen2.5-VL-3B} and \textit{Qwen2.5-VL-7B}. We use \textit{Qwen3-VL-30B-Thinking} as the verifier across all experiments, as its strong reasoning capabilities provide a robust reward signal.

\noindent \textbf{Baselines.} We compare our approach against competitive baselines covering a broad range of techniques for failure mode discovery. Since no prior work tailors the failure-discovery process to a specific target VLM and dataset combination, we implement all baselines from scratch (except ConMe~\cite{huang2024conme}):

\begin{enumerate}
    \item \textbf{Zero Shot:} We directly prompt the untrained questioner without any fine-tuning or adaptation.

    \item \textbf{ConMe:} We follow the original in-context learning and prompting strategy from~\cite{huang2024conme}, which employs a two-turn process where second-turn queries are conditioned on the VLM's initial responses to expose reasoning failures.

    \item \textbf{Supervised Fine-Tuning (SFT):} We fine-tune the questioner on questions drawn from existing failure mode discovery benchmarks~\cite{fu2024blink,wu2024v,li2025core,tong2024eyes,yue2025mmmu,fu2024ocrbench,masry2025chartqapro} before prompting it to generate questions.

    \item \textbf{Expert Iteration:} We iteratively fine-tune the questioner for $5$ rounds. In each round, the questioner generates questions that are filtered via rejection sampling using the same verifier as our RL method. The retained failure cases are then used to fine-tune the questioner for the next round, progressively tailoring generation to the vulnerabilities of the target VLM.

    \item \textbf{RL+SFT:} We first fine-tune the questioner on existing failure mode datasets, then train it using our RL framework.
\end{enumerate}

\noindent \textbf{Evaluation Metrics.} We evaluate the failure discovery process along two dimensions: \textit{potency} (how effectively a method elicits failures) and \textit{diversity} (how varied the discovered failures are). All metrics are computed using only valid questions. Our metrics are:

\begin{enumerate}
    \item \textbf{Question Validity Rate (QVR):} The fraction of generated questions that satisfy the validity criteria defined in Section~\ref{sec:formulation}.

    \item \textbf{Failure Discovery Rate (FDR):} The fraction of valid questions that are incorrectly answered by the target VLM. This is the core metric for measuring a method's potency.

    \item \textbf{Semantic Diversity (SD):} Computed using the Vendi Score~\cite{friedman2023vendi}, which quantifies the effective diversity of generated questions based on their embeddings. This metric is more robust than pairwise cosine similarity-based alternatives.

    \item \textbf{Lexical Diversity (LD):} The entropy of the question prefix distribution (first two words), which measures the diversity of question types (e.g., ``How many'' vs.\ ``What is''). Higher entropy indicates a richer mixture of question formats.

    \item \textbf{Skill Coverage (SC):} The average number of distinct cognitive skills required to answer a generated question, serving as a proxy for question complexity.

    \item \textbf{Number of Skills (\# Skills):} The total number of distinct skills probed by a method's generated questions. We filter out skills comprising $20$ or fewer questions to ensure this metric is not inflated by outliers.
\end{enumerate}

\noindent \textbf{Evaluation Model.} We use Gemini 3 Flash to compute QVR and FDR scores, as it is independent of the training pipeline and provides an unbiased assessment.

\noindent \textbf{Human Validation.} To verify the reliability of automatic evaluation, we collect human judgments for both QVR and FDR via Amazon Mechanical Turk, with five annotations per example. Results indicate strong agreement between human and Gemini judgments: $80\%$ agreement on QVR ($2000$ samples) and $75\%$ on FDR ($1200$ samples), confirming that Gemini serves as a reliable proxy for human evaluation (see Appendix for details).

\subsection{Does RL Help in Failure Discovery?}
\label{sec:main_results}

Table~\ref{tab:evaluation_metrics} compares our RL framework against the baselines across all evaluation axes. We organize our analysis around three aspects: question validity, potency of elicited failures, and diversity of discovered failures.

\begin{table}[h]
\centering
\caption{Evaluation metrics for different methods using \textit{Qwen-2.5-VL-3B} as both questioner and answerer. \textbf{Best} and \underline{second best} results are highlighted. Higher is better for all metrics.}
\label{tab:evaluation_metrics}
\small
\resizebox{0.85\linewidth}{!}{%
\begin{tabularx}{\linewidth}{l *{6}{>{\centering\arraybackslash}X}}
\toprule
\textbf{Method} & \textbf{QVR(\%)} & \textbf{FDR(\%)} & \textbf{SD} & \textbf{LD} & \textbf{\# Skills} & \textbf{Skill Cov.} \\
\midrule
Zero Shot         & 83.73             & 32.20             & 56.27             & 0.30              &  \underline{109}              & 3.06 \\
SFT               & 83.86             & 37.02             & 43.80             & 0.23              &  99              & 3.13  \\
Expert Iter.      & \textbf{88.37}    & 38.36             & 53.42             & 0.19              &  103              & 3.10 \\
ConMe             & 76.63             & 39.89             & 61.78             & 0.30              &  107              & 3.15  \\
\rowcolor{gray!15}RL                & 83.44             & \underline{47.58} & \underline{64.96} & \underline{0.53}    & \textbf{126}    & \textbf{3.30} \\
\rowcolor{gray!15}RL+SFT            & \underline{86.70}             & \textbf{50.73}    & \textbf{85.41}    & \textbf{0.57} &  \underline{109} & \underline{3.27} \\
\bottomrule
\end{tabularx}
}
\end{table}

\noindent \textbf{Question Validity.} All methods achieve comparable Question Validity Rates (QVR). The base RL model shows a marginal decrease in QVR due to the inherent trade-off between failure elicitation, diversity, and validity (analyzed in Section~\ref{sec:ablations}). This is mitigated by applying RL to a post-SFT model (RL+SFT), which recovers a QVR of $86.70\%$.

\noindent \textbf{Potency of Elicited Failures:} Our RL-based approaches achieve substantially higher Failure Discovery Rates, $47.58\%$ for RL and $50.73\%$ for RL+SFT, compared to the strongest non-RL baseline, ConMe, at $39.89\%$. Our human study (see Appendix) confirms that these FDR gains reflect genuine failures rather than artifacts of the automated evaluator.

\begin{table}[h!]
  \centering
  \caption{Average failure rate per skill across methods, computed as the per-skill mean over clusters with more than 100 occurrences.}
  \label{tab:method_performance_per_skill}
  \begin{tabular}{lc}
    \toprule
    \textbf{Method} & \textbf{Failure rate per skill (\%)} \\
    \midrule
    ZeroShot         & 33.38 \\
    SFT              & 34.72 \\
    ExpertIteration  & 38.54 \\
    ConMe            & 40.36 \\
    RL               & 48.27 \\
    RL + SFT         & \textbf{48.82} \\
    \bottomrule
  \end{tabular}
\end{table}

To verify that these improvements do not arise from over-exploiting a narrow set of failure modes, we analyze performance across individual skills in Table \ref{tab:method_performance_per_skill}. Our RL method maintains the highest average per-skill failure rate ($48.27\%$) while exploring $126$ unique skills. Moreover, RL-generated questions exhibit higher Skill Coverage, requiring $3.30$ simultaneous skills per query compared to $3.15$ for ConMe, the best baseline. These results demonstrate that RL effectively explores the VLM's failure landscape without sacrificing discovery depth, particularly when less constrained by prior SFT distributions.

\noindent \textbf{Diversity of Discovered Failures:} RL-based methods substantially increase the variety of discovered failures along both semantic and lexical dimensions. As shown in Table~\ref{tab:evaluation_metrics}, RL achieves a Semantic Diversity of $64.96$ and RL+SFT reaches $85.41$, compared to $61.78$ for ConMe, the next best baseline. The most pronounced difference is in Lexical Diversity: RL methods achieve an LD of $0.53$ and $0.57$, nearly doubling ConMe's $0.30$. This indicates that the RL agent generates a rich variety of question types (\emph{Is}, \emph{Does}, \emph{How many}, \emph{What}, \emph{Where}, etc.) rather than relying on a narrow set of templates. Furthermore, RL discovers $126$ distinct skills, surpassing all baselines which plateau at $99$--$109$ skills.

\begin{figure}[h]
    \centering
    \includegraphics[width=0.98\linewidth]{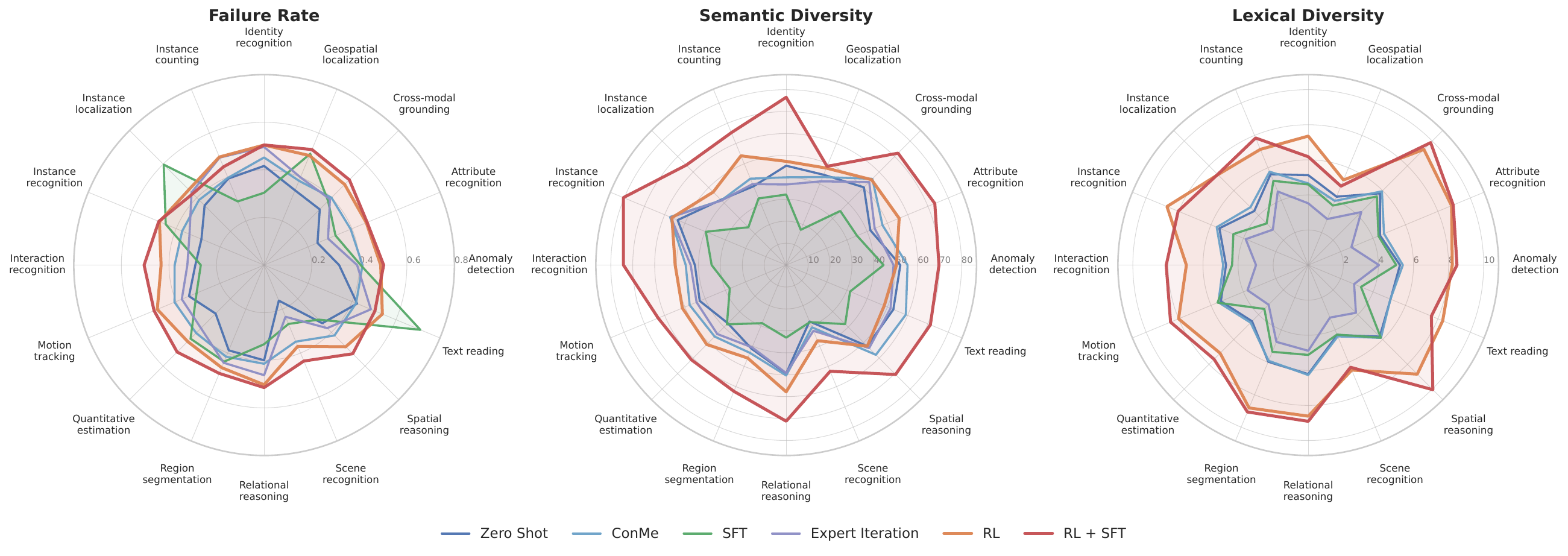}
    \caption{Per-meta-skill comparison of Failure Rate, Semantic Diversity, and Lexical Diversity across methods. RL-based methods consistently cover a larger area, indicating broader and deeper failure discovery across the skill space.}
    \label{fig:combined_radar}
\end{figure}

The aggregate metrics in Tables~\ref{tab:evaluation_metrics} and~\ref{tab:method_performance_per_skill} summarise each method with a single value, but this can obscure an important distinction: a method may achieve high averages by excelling on a few skills while neglecting others, rather than consistently discovering failures across the full skill space. True failure discovery demands both breadth, covering many skills, and depth, achieving high FDR and diversity within each. To evaluate this, we disaggregate FDR, Semantic Diversity, and Lexical Diversity by meta-skill and compare methods using radar plots (Figure~\ref{fig:combined_radar}). RL-based methods consistently span a larger area than all baselines across meta-skills, confirming that their improvements are uniformly distributed rather than concentrated in a narrow subset of failure modes.

\noindent \textbf{Summary:} RL-based methods demonstrate superior performance across all evaluation axes by jointly optimizing for failure discovery and diversity. These results confirm that the multi-objective reward signal is essential for uncovering a broad spectrum of genuine VLM vulnerabilities that remain hidden from static and iterative methods.

\subsection{Are the Discovered Failures Novel?}
\label{sec:novelty}

To assess the novelty of failures identified by our framework, we examine two dimensions: the breadth of skills discovered and the compositional complexity of generated questions.

\begin{figure}[h]
    \centering
    \includegraphics[width=0.6\linewidth]{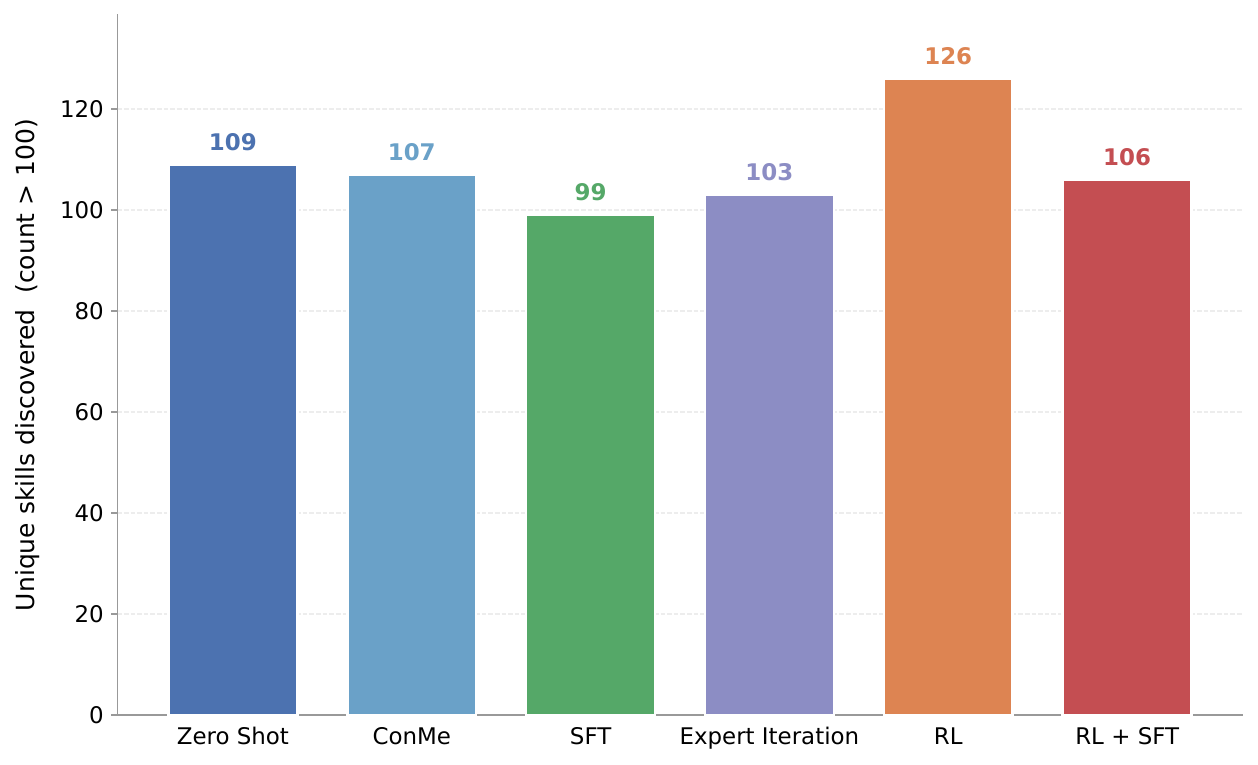}
    \caption{Number of unique skills explored by each method.}
    \label{fig:skill_distribution_by_method}
\end{figure}

\noindent \textbf{Breadth of Discovered Skills.} Figure~\ref{fig:skill_distribution_by_method} compares the number of skills explored by each method. Our RL method covers a substantially wider range of skills ($126$) than the baselines (at most $109$ for ZeroShot, with $99$ for SFT and $107$ for ConMe). Beyond covering more skills overall, RL-based methods also discover skills that no baseline explores, we analyze these exclusive discoveries in detail below.

\begin{figure}[h]
  \centering
   \begin{subfigure}[b]{0.48\textwidth}
    \includegraphics[width=\textwidth]{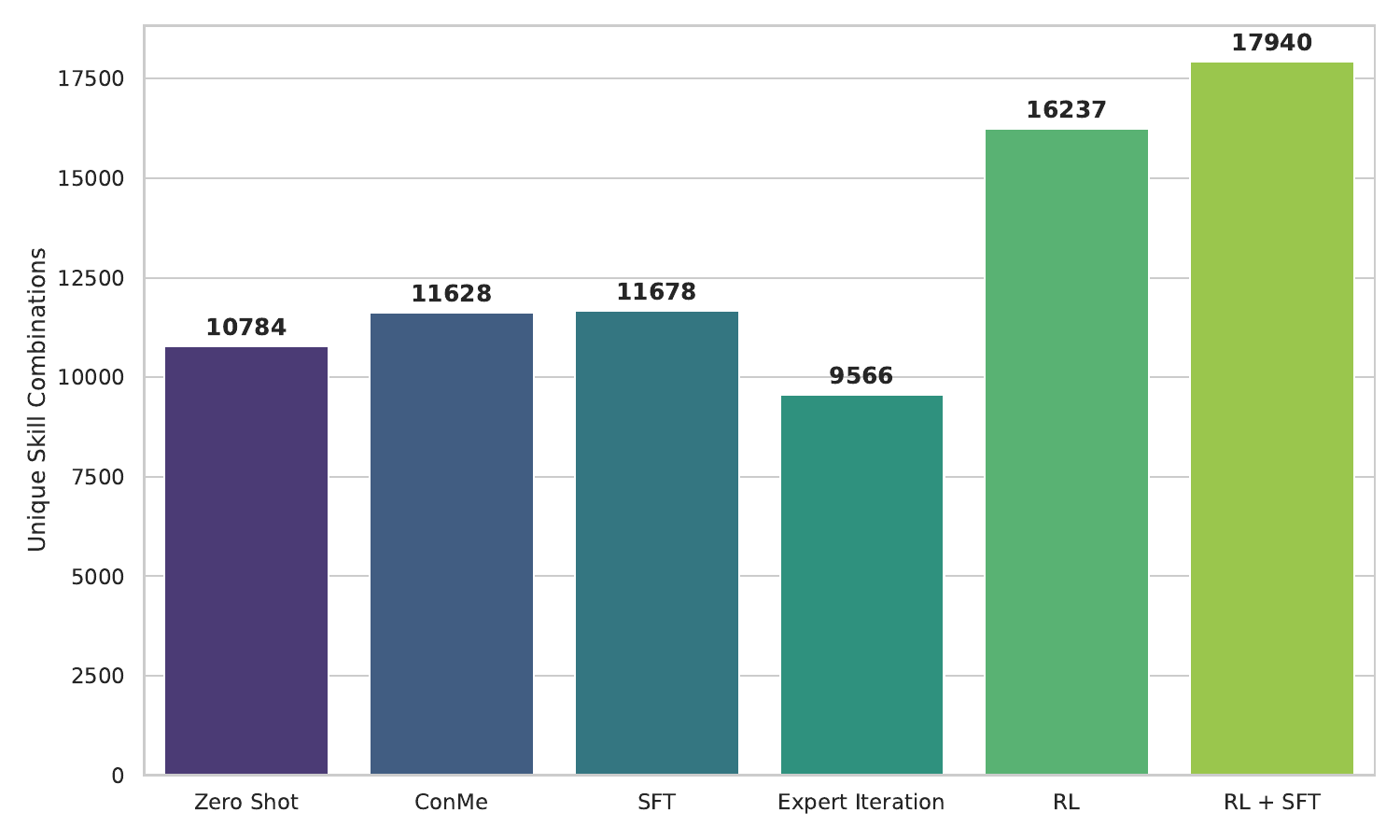}
    \caption{Unique skill combinations explored by each method.}
    \label{fig:unique_skill_combinations}
  \end{subfigure}
   \hfill
  \begin{subfigure}[b]{0.48\textwidth}
    \includegraphics[width=\textwidth]{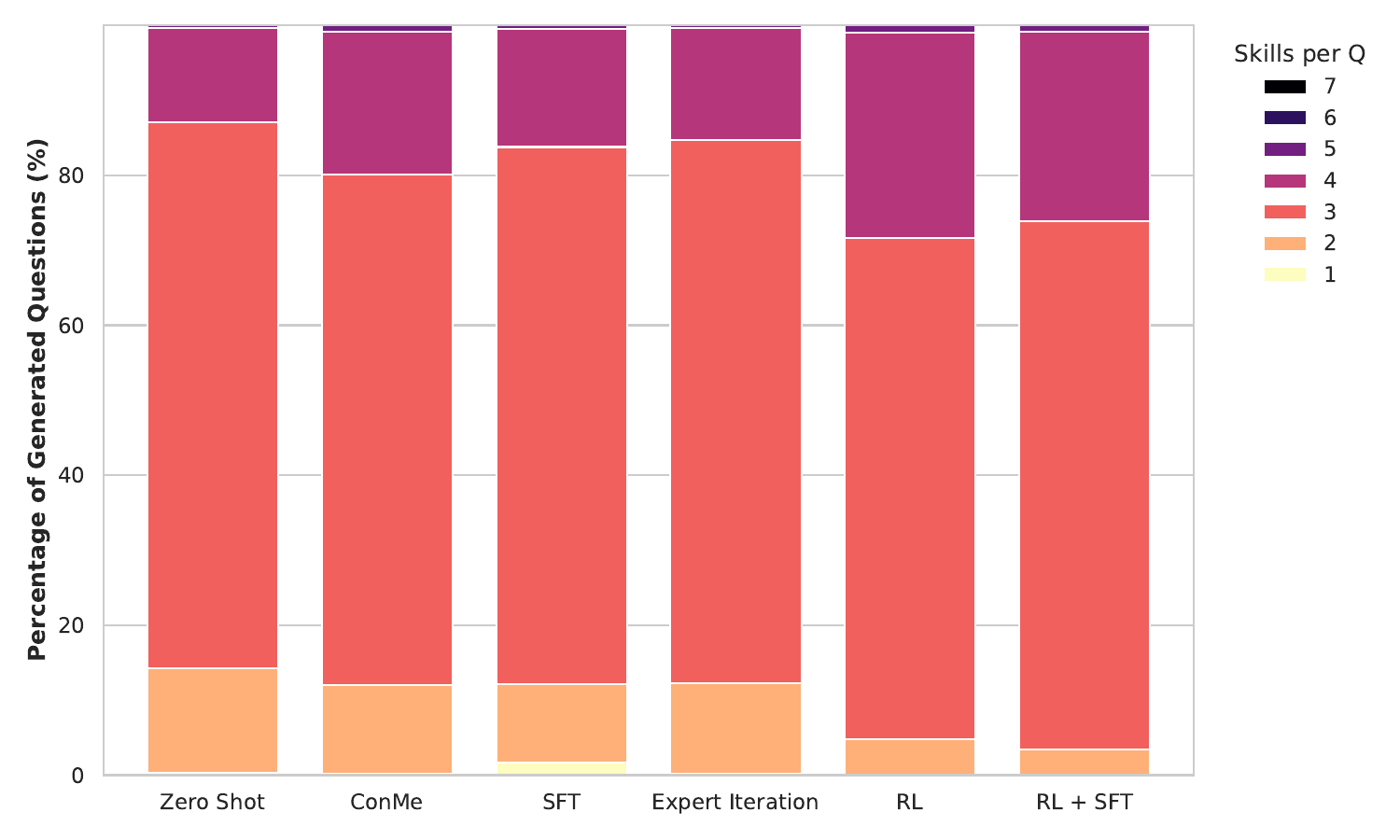}
    \caption{Distribution of question complexity across methods based on skill counts.}
    \label{fig:question_complexity_distribution}
  \end{subfigure}
  \caption{Analysis of skill diversity and question complexity across methods.}
\end{figure}

\begin{table}[h!]
\centering
    \small
    \caption{Exclusive skills per method group with per-skill failure rates. Skills in the top section are discovered only by RL methods (11 total); skills in the bottom section are discovered only by baselines (6 total).}
    \label{tab:exclusive_skills}
    \begin{tabular}{lcc}
    \toprule
      \textbf{Skill} & \textbf{RL (\%)} & \textbf{RL+SFT (\%)} \\
      \midrule
      \multicolumn{3}{l}{\textit{Exclusive to RL methods (avg $\sim$50.4\%)}} \\
      \midrule
      hierarchical reasoning      & 61.2 & 53.8 \\
      pattern matching            & 56.4 & 48.6 \\
      part comparison             & 55.7 & 48.3 \\
      connector connectivity      & 54.6 & 50.4 \\
      trajectory reasoning        & --   & 54.2 \\
      attachment detection        & 52.3 & 46.9 \\
      adjacency relations         & 51.4 & 41.8 \\
      damage detection            & 51.0 & 47.9 \\
      continuity detection        & --   & 48.0 \\
      anatomical localization     & 45.5 & 47.8 \\
      roadway recognition         & 44.8 & --   \\
      \midrule
      \textbf{Skill} & \multicolumn{2}{c}{\textbf{Baseline avg (\%)}} \\
      \midrule
      \multicolumn{3}{l}{\textit{Missed by both RL methods (avg $\sim$27.6\%)}} \\
      \midrule
      image-caption matching      & \multicolumn{2}{c}{37.4} \\
      ui recognition              & \multicolumn{2}{c}{35.9} \\
      criteria-based selection    & \multicolumn{2}{c}{27.8} \\
      waveform recognition        & \multicolumn{2}{c}{26.4} \\
      organism health inference   & \multicolumn{2}{c}{21.4} \\
      cloud detection             & \multicolumn{2}{c}{16.8} \\
      \bottomrule
      \end{tabular}
\end{table}

\noindent \textbf{Exclusive Skills.} Table~\ref{tab:exclusive_skills} lists the exclusive skills discovered by RL-based methods and baselines. RL methods discover $11$ exclusive skills that no baseline discovers. RL surfaces exclusive failures spanning reasoning, detection, localization, and recognition meta-skills such as  hierarchical reasoning ($61.2\%$), damage detection ($51.0\%$), and connector connectivity ($54.6\%$). These skills exhibit an average per-skill failure rate of ${\sim}50.4\%$, substantially exceeding RL's overall failure rate of $47.58\%$, indicating these are challenging skills for target VLM that the RL agent learns to target. In contrast, the $6$ skills exclusive to baselines exhibit a markedly lower average failure rate of ${\sim}27.6\%$, suggesting that they correspond to relatively mild weaknesses. The disparity in both the number and severity of exclusive skills reinforces that RL-driven exploration preferentially surfaces hard, structurally complex failure modes compared to static methods.

\noindent \textbf{Compositional Complexity.} Beyond discovering more skills, RL-generated questions combine skills in more diverse ways. Figure~\ref{fig:unique_skill_combinations} measures the total number of unique skill combinations across all questions generated by each method. RL discovers $4,559$ more unique combinations than the next best baseline SFT, indicating that RL explores a far richer space of multi-skill queries. Figure~\ref{fig:question_complexity_distribution} further shows that RL produces a higher proportion of questions requiring four or more simultaneous skills, confirming that RL-generated questions demand proficiency across multiple cognitive dimensions. For example, instead of simple identification questions, the RL questioner generates compositionally complex queries such as: \textit{``What traffic light is located to the immediate left of the train and south-west of the stop sign above it?''} Answering this question requires multi-hop reasoning: first locating all traffic lights to the immediate left of the train, then identifying the stop sign, and finally selecting the traffic light south-west of it. Such queries simultaneously probe spatial reasoning, object localization, and relational understanding.

The emergence of these exclusive skills and multi-hop queries demonstrates that our framework identifies a more nuanced class of failures that test fine-grained understanding of the target VLM, going beyond the capabilities of static probing methods.

\subsection{How Does Failure Discovery Evolve During Training?}
\label{sec:training_dynamics}

\begin{figure}[h!]
    \centering
    \includegraphics[width=0.8\textwidth]{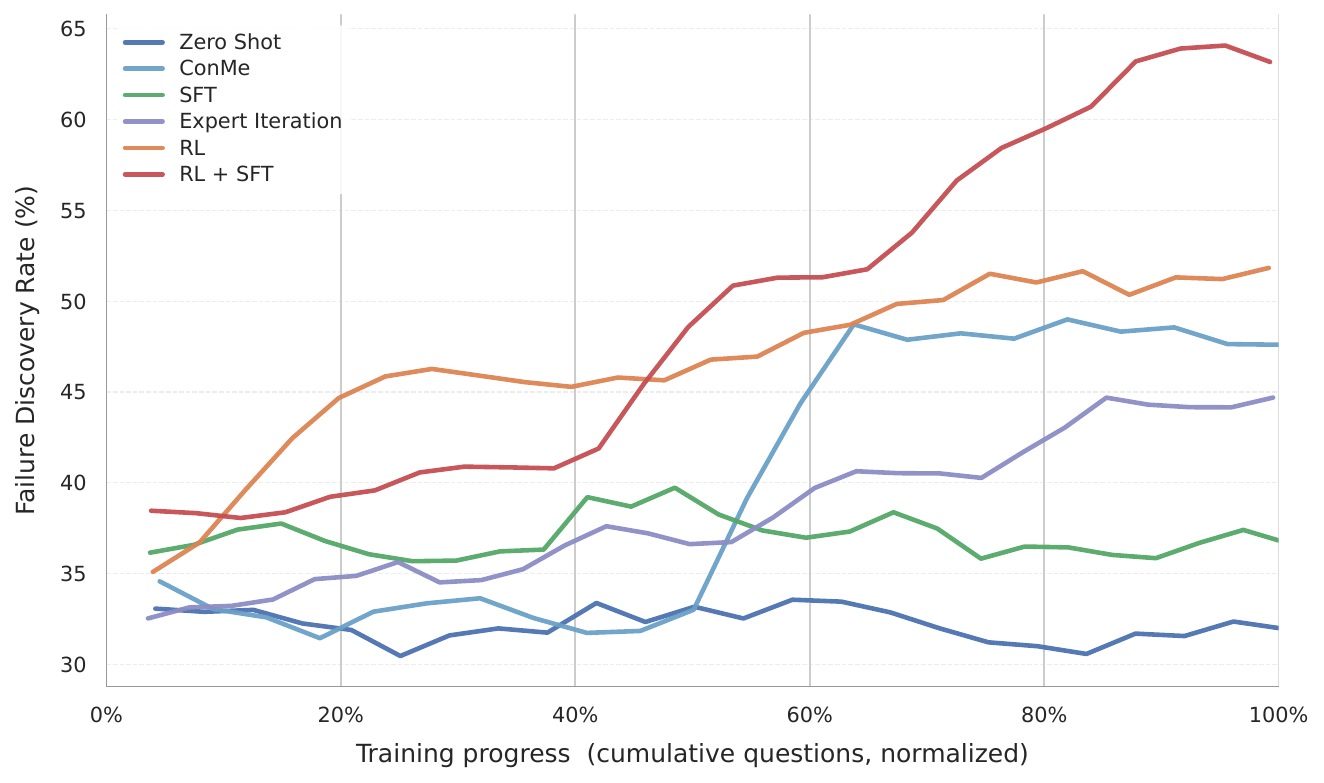}
    \caption{Comparison of cumulative failure discovery rates across different methods. Our RL-based approach shows a consistent increase in unique failure detection over training iterations.}
    \label{fig:cumulative_failure_rate}
\end{figure}

To understand the dynamics underlying our framework, we analyze how failure-discovery behavior evolves over the course of RL optimization. We examine two aspects: the evolution of the Failure Discovery Rate and the temporal dynamics of meta-skills.

\noindent \textbf{Evolution of Failure Discovery Rate.} Figure~\ref{fig:cumulative_failure_rate} plots FDR as a function of cumulative questions generated. Static methods (Zero-Shot, ConMe, and SFT) exhibit a flat trend, as their probing strategies are fixed at inference time. In contrast, methods that employ iterative optimization, Expert Iteration and RL, show a consistent increase in FDR as training progresses, with RL exhibiting the most pronounced improvement. This confirms that the RL questioner progressively adapts its probing strategy to exploit the target VLM's weaknesses.

\begin{figure}[h!]
    \centering
    \includegraphics[width=\linewidth]{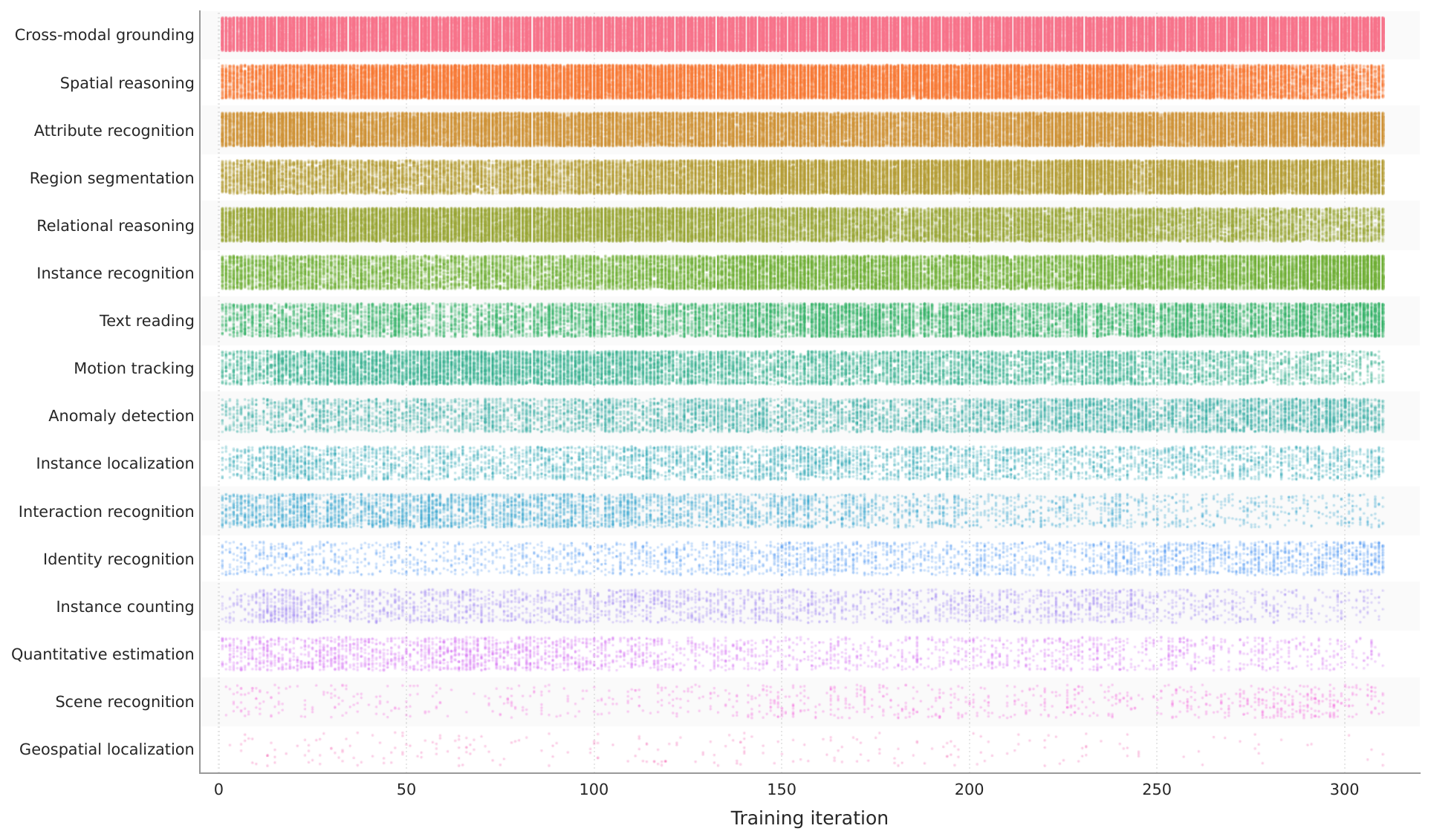}
    \caption{Meta-skill density during RL training. Dense, continuous bands indicate sustained probing of discovered skills; sparse or fading bands indicate skills that the agent moves away from over time.}
    \label{fig:skill_emergence_density}
\end{figure}

\noindent \textbf{Temporal Dynamics of Meta-Skills.} Beyond the overall rise in FDR, we investigate how individual meta-skills are discovered and exploited over the course of training. Figure~\ref{fig:skill_emergence_density} visualizes the density of meta-skill probing across training steps. We observe three distinct temporal patterns:

\begin{itemize}
    \item \textit{Persistent skills}, such as ``Spatial Reasoning'' and ``Cross-modal Grounding'', remain consistently vulnerable throughout optimization and are probed at a steady rate;
    \item \textit{Early-peaking skills}, such as ``Quantitative Estimation'' and ``Interaction Recognition'', which the agent exploits heavily in early training before their prevalence diminishes;
    \item \textit{Late-emerging skills}, such as ``Identity Recognition'' and ``Anomaly Detection'', which are only discovered as training matures and the agent exhausts more obvious failure modes.
\end{itemize}

The varying rates at which these meta-skills emerge and recede suggest that the RL agent follows an implicit curriculum: it first prioritizes meta-skills that consistently yield failures, then progressively pivots toward more elusive failure modes as the easier ones are exhausted. This qualitative refinement of the search strategy enables the RL framework to uncover a diverse landscape of VLM failures that lies beyond the reach of static methods.

\begin{figure}[h!]
    \centering
    \includegraphics[width=0.95\textwidth]{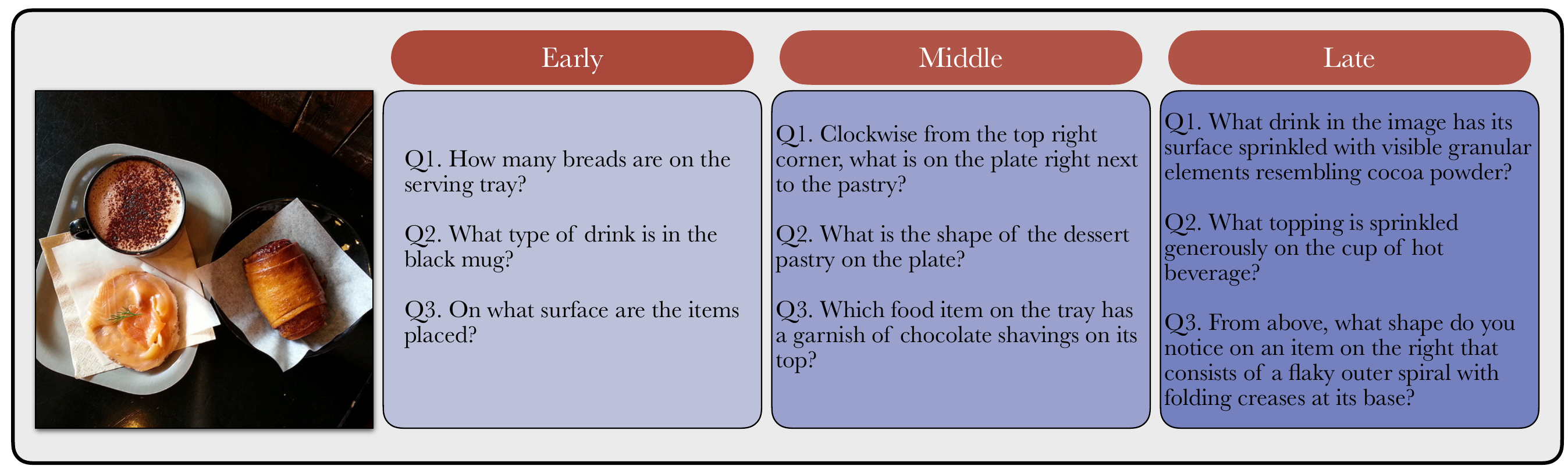}
    \caption{Evolution of RL-generated questions for a static image at different stages of training. The generated questions increase in complexity by probing fine-grained reasoning that requires proficiency in multiple skills.}
    \label{fig:question_evolution}
\end{figure}

\noindent \textbf{Qualitative Evolution of Generated Questions.}
The trends above are accompanied by a visible shift in the structure and complexity of the questions produced by the RL agent. Figure~\ref{fig:question_evolution} illustrates this progression for a single image. During training, the generated questions evolve along different axes, such as the specificity of the reference, linguistic complexity, and visual granularity. Early questions are short and direct as they explicitly refer to objects in the image. Middle questions specify referred objects through spatial reasoning and relative positioning, while questions towards the end describe them through their visual attributes, e.g., ``surface sprinkled with granular elements''. The VLM must recognise fine-grained elements to recognise which objects are being referred to. Similarly, linguistic complexity increases over the course of training as the generated questions get longer and more descriptive. This progression reflects a shift from single-hop questions to multi-hop reasoning questions in which the VLM must first resolve an indirect perceptual description before answering the actual query. Finally, the questions also evolve in the visual granularity they target. Early questions probe coarse-grained, whole-object properties such as counting or object category, while later questions demand attention to fine-grained attributes such as surface texture, shape geometry, and material appearance. This qualitative progression suggests that the RL agent not only discovers \emph{what} to probe but also learns \emph{how} to probe it more effectively, generating increasingly targeted queries that surface failure modes inaccessible to simpler probing strategies.

\subsection{What Drives Effective Failure Discovery?}
\label{sec:ablations}

To understand the contribution of each design choice in our framework, we systematically ablate the reward components and key hyperparameters. All ablations use \textit{Qwen2.5-VL-3B} as both questioner and candidate VLM.

\subsubsection{Reward Component Analysis}

Table~\ref{tab:ablation_reward} presents the impact of each component in Equation~\ref{eq:final_reward}. We highlight three key findings.

\begin{table}[h]
\centering
\caption{Ablation study of reward components. We evaluate the contribution of Semantic Diversity (Sem.), Lexical Diversity (Lex.), and Invalid Penalty (Pen.). Row~4 (shaded) is our full reward.}
\label{tab:ablation_reward}
\small
\begin{tabularx}{0.9\linewidth}{@{}ccc *{5}{>{\centering\arraybackslash}X}@{}}
\toprule
\multicolumn{3}{c}{\textbf{Reward Components}} & \multicolumn{5}{c}{\textbf{Metrics}} \\
\cmidrule(r){1-3} \cmidrule(l){4-8}
\textbf{Sem.} & \textbf{Lex.} & \textbf{Pen.} & \textbf{QVR} & \textbf{FDR} & \textbf{SD} & \textbf{LD} & \textbf{SC} \\ \midrule
\checkmark & \checkmark & -- & 75.82 & 47.47 & 63.25 & \textbf{0.56} & \underline{123} \\
-- & -- & \checkmark & \textbf{89.67} & \textbf{66.70} & 59.29 & 0.16 & 108 \\
\checkmark & -- & \checkmark & \underline{87.18} & \underline{56.54} & \textbf{68.48} & 0.19 & 118 \\
\rowcolor{gray!15} \checkmark & \checkmark & \checkmark & 83.44 & 47.58 & \underline{64.96} & \underline{0.53} & \textbf{126} \\ \bottomrule
\end{tabularx}
\end{table}

\noindent \textbf{Invalid penalty is essential for question quality.} Removing the invalid penalty $P_Q$ (Row~1) causes QVR to drop from $83.44\%$ to $75.82\%$, confirming that this component is critical for directing the questioner toward valid, well-formed queries.

\noindent \textbf{Lexical Diversity prevents reward hacking.} Omitting the Lexical Diversity component (Rows~2 and~3) leads to a collapse in LD, which falls as low as $0.16$. This indicates that the agent resorts to reward hacking, repeatedly using a narrow set of question templates (e.g., exclusively asking ``How many...'') to exploit known weaknesses. Although Row~2 achieves the highest FDR of $66.70\%$, this reflects repetitive exploitation of a small number of failure modes rather than broad discovery. Incorporating Lexical Diversity (Row~4) penalizes prefix repetition, which moderates FDR to $47.58\%$ but forces the agent to explore a far wider range of visual vulnerabilities.

\noindent \textbf{Semantic Diversity broadens the failure landscape.} Comparing Rows~2 and~3, adding Semantic Diversity increases SD from $59.29$ to $68.48$ and increases the number of skills discovered ($108 \rightarrow 118$), confirming its role in pushing the agent beyond surface-level exploitation.

\subsubsection{Scaling number of Questions per Image}

We investigate the effect of scaling the number of questions generated per image ($Qs$) during training. As shown in Table~\ref{tab:ablation_questions}, $Qs=2$ achieves the best balance between potency and breadth, maintaining a high FDR ($47.58\%$) while increasing the number of unique skills from $78$ to $126$.

\begin{table}[h!]
\centering
\caption{Ablation on queries per image ($Qs$).}
\label{tab:ablation_questions}
\small
\setlength{\tabcolsep}{3pt}
\begin{tabular}{@{}c ccccc@{}}
\toprule
\textbf{\# Qs} & \textbf{QVR} & \textbf{FDR} & \textbf{SD} & \textbf{LD} & \textbf{Skills} \\ \midrule
1 & 83.02 & \textbf{49.41} & 67.69 & 0.52 & 78 \\
2 & \textbf{83.44} & 47.58 & 64.96 & 0.53 & 126 \\
4 & 79.52 & 42.79 & \textbf{73.57} & \textbf{0.53} & \textbf{137} \\ \bottomrule
\end{tabular}
\end{table}

At $Qs=4$, discovery breadth reaches a maximum of $137$ skills, but FDR drops to $42.79\%$. This trade-off suggests that once the most obvious vulnerabilities have been identified through the first few questions, the questioner must target increasingly subtle visual details to elicit further errors. Despite the lower per-query success rate, the superior Semantic Diversity ($73.57$) and skill count at $Qs=4$ demonstrate that multi-query generation is essential for uncovering the long tail of sophisticated VLM failures.

\subsubsection{Impact of Prefix Length on Lexical Diversity}

We evaluate the sensitivity of the Lexical Diversity reward ($\delta_{\text{ifreq}}$) to the prefix length $L$, which defines the granularity of a question ``type.'' As shown in Table~\ref{tab:ablation_prefix_length}, using a single token ($L=1$) results in the lowest LD ($0.16$), indicating template collapse: the agent merely swaps initial words (e.g., ``What'' vs.\ ``How'') to satisfy the reward without meaningfully changing the query structure.

\begin{table}[h!]
  \centering
  \caption{Ablation on question prefix length ($L$) for the Lexical Diversity reward.}
  \label{tab:ablation_prefix_length}
  \small
  \setlength{\tabcolsep}{4pt}
  \begin{tabular}{@{}c ccccc@{}}
    \toprule
    \textbf{L} & \textbf{QVR} & \textbf{FDR} & \textbf{SD} & \textbf{LD} & \textbf{Skills} \\ \midrule
    1 & \textbf{87.52} & \textbf{63.98} & 65.73 & 0.16 & 113 \\
    2 & 83.44 & 47.58 & 64.96 & \textbf{0.53} & \textbf{126} \\
    3 & 84.56 & 45.62 & \textbf{82.71} & 0.45 & 109 \\
    4 & 86.00 & 53.82 & 74.97 & 0.32 & 108 \\ \bottomrule
  \end{tabular}
\end{table}

Setting $L=2$ provides the most effective regularization, achieving a peak LD of $0.53$ by capturing distinct question intents (e.g., ``How many'' vs.\ ``What color''). At $L=4$, both LD and the number of unique skills decrease, suggesting that an overly specific prefix definition over-constrains the questioner and produces a sparse reward signal that is difficult to optimize against.

\subsection{How Does Failure Discovery Scale with Model Capacity?}

\begin{table}[ht!]
\centering
\caption{Cross-model evaluation across different questioner and answerer models. SD denotes Semantic Diversity and LD denotes Lexical Diversity. Higher is better for all metrics.}
\label{tab:cross_model}
\small
\setlength{\tabcolsep}{5pt}
\renewcommand{\arraystretch}{1.2}
\resizebox{\linewidth}{!}{%
\begin{tabular}{lllccccc}
\toprule
\textbf{Questioner} & \textbf{Method} & \textbf{Answerer} & \textbf{QVR (\%)} & \textbf{FDR (\%)} & \textbf{SD} & \textbf{LD} & \textbf{\# Skills} \\
\midrule
\multirow{6}{*}{Qwen-2.5-VL-3B} & \multirow{3}{*}{RL} & Qwen-2.5-VL-3B & 83.44 & 47.58 & 64.96 & 0.53 & 126 \\
 & & Qwen-2.5-VL-7B & 81.94 & 44.55 & 48.23 & 0.54 & 103 \\
 & & Qwen-3-VL-8B & 85.11 & 29.37 & 59.88 & 0.44 & 116 \\
\cmidrule{2-8}
 & \multirow{3}{*}{Exp. Iter.} & Qwen-2.5-VL-3B & 88.37 & 38.36 & 53.42 & 0.19 & 103 \\
 & & Qwen-2.5-VL-7B & 87.97 & 28.73 & 53.90 & 0.19 & 103 \\
 & & Qwen-3-VL-8B & 87.64 & 23.67 & 51.75 & 0.18 & 101 \\
\midrule
\multirow{6}{*}{Qwen-2.5-VL-7B} & \multirow{3}{*}{RL} & Qwen-2.5-VL-3B & 79.70 & 51.33 & 75.16 & 0.56 & 121 \\
 & & Qwen-2.5-VL-7B & 81.65 & 54.25 & 55.71 & 0.57 & 96 \\
 & & Qwen-3-VL-8B & 84.55 & 38.97 & 54.86 & 0.48 & 104 \\
\cmidrule{2-8}
 & \multirow{3}{*}{Exp. Iter.} & Qwen-2.5-VL-3B & 85.47 & 40.69 & 61.95 & 0.22 & 101 \\
 & & Qwen-2.5-VL-7B & 84.87 & 31.46 & 62.97 & 0.22 & 107 \\
 & & Qwen-3-VL-8B & 85.39 & 26.76 & 61.76 & 0.22 & 105 \\
\bottomrule
\end{tabular}
}
\end{table}

Table~\ref{tab:cross_model} shows the results for different combinations of questioner and answerer models. These results use Gemini as the verifier. We analyze how questioner capacity, answerer capacity, and training method interact to influence failure discovery.

As expected, increasing the answerer model's capacity while keeping the questioner fixed leads to a consistent drop in FDR. Instead, scaling the questioner from 3B to 7B yields steady FDR gains across all answerer models and both training methods. Under the RL-based method, switching from 3B to 7B questioner results in relative improvement of $\sim9\%$, $\sim21\%$ and $\sim32\%$ for 3B, 7B and 8B answerer, respectively. Interestingly, the largest relative improvement occurs with the strongest 8B answerer model, suggesting that question capacity is an important axis for failure discovery.

Furthermore, the table shows a widening gap between RL and Expert Iteration as the answerer model's capacity increases. Expert iteration shows diminishing relative gains as the questioner capacity increases, compared to the RL method, as the relative improvement from a 3B questioner to a 7B questioner with an 8B answerer is just $13\%$, compared to $32\%$ with RL for the same setting. The lexical diversity remains low for expert iteration across model combinations, suggesting a narrow set of repetitive question templates that strong answerer models readily overcome. In contrast, RL's diversity-driven exploration produces more varied probing strategies that continue to expose failures even in more capable models.

\subsection{Which Failures Persist Across Model Scales?}
\label{sec:failure_landscape}

We examine the failure landscape across different scales of questioner and answerer models. For each questioner, we show the universally hard skills that all answerer models struggle with. Tables~\ref{tab:landscape_3b} and~\ref{tab:landscape_7b} present the top-10 universally hard and most model-specific skills for the 3B and 7B questioners, respectively.

\begin{table}[h!]
\centering
\caption{Failure landscape for the 3B questioner across answerer scales. \textbf{Left:} Universally hard skills ranked by minimum FDR across all answerers, indicating failures that persist regardless of model capacity. \textbf{Right:} Model-specific skills ranked by FDR standard deviation, indicating failures that are resolved with increased scale. FDR values are percentages.}
\label{tab:landscape_3b}
\resizebox{\textwidth}{!}{%
\begin{tabular}{lcccc||lccccc}
\toprule
\multicolumn{4}{c}{\textbf{Universally Hard Skills}} & & \multicolumn{5}{c}{\textbf{Model-Specific Skills}} \\
\cmidrule(lr){1-4} \cmidrule(lr){6-10}
\textbf{Skill} & \textbf{3B} & \textbf{7B} & \textbf{8B} & & \textbf{Skill} & \textbf{3B} & \textbf{7B} & \textbf{8B} & $\boldsymbol{\sigma}$ \\
\midrule
Sign Recognition              & 57.5 & 49.2 & 49.9 & & Fine-Grained Diff.\ Detection & 58.2 & 44.4 & 12.2 & 23.6 \\
Logo Recognition              & 56.6 & 48.6 & 49.3 & & UI Recognition                & 56.6 & 60.9 & 23.5 & 20.4 \\
Label Detection               & 61.9 & 49.4 & 43.1 & & Pattern Matching              & 56.4 & 51.6 & 21.2 & 19.1 \\
Connector Connectivity        & 54.6 & 48.2 & 41.3 & & Adjacency Relations           & 51.4 & 46.3 & 16.4 & 18.9 \\
Analogy Interpretation        & 55.3 & 50.5 & 40.9 & & Roadway Recognition           & 44.8 & 52.4 & 17.6 & 18.3 \\
Traffic Sign Recognition      & 59.6 & 48.6 & 40.7 & & Image Completion              & 42.9 & 51.7 & 16.7 & 18.2 \\
OCR                           & 55.0 & 41.6 & 40.6 & & Number Sequence Recognition   & 65.3 & 40.0 & 30.2 & 18.1 \\
Material Identification       & 55.7 & 41.4 & 39.5 & & Part Comparison               & 55.7 & 30.4 & 22.2 & 17.4 \\
Electronic Device Recognition & 53.4 & 47.5 & 39.2 & & Container Content Recognition & 57.4 & 42.9 & 23.2 & 17.1 \\
Identity Verification         & 52.3 & 38.6 & 40.2 & & Shadow Separation             & 48.4 & 32.5 & 15.1 & 16.6 \\
\bottomrule
\end{tabular}%
}
\end{table}

\noindent \textbf{Universally Hard Skills.} Across both questioner scales, a consistent set of skills emerges as challenging for all answerers regardless of capacity. Skills such as \textit{Sign Recognition}, \textit{Logo Recognition}, \textit{Label Detection}, and \textit{Traffic Sign Recognition} maintain high failure rates even for the largest 8B answerer. The persistence of these failures across scales suggests they reflect limitations that increased model capacity alone does not mitigate. Moreover, the 7B questioner surfaces consistently higher FDRs compared to the 3B questioner (min FDR of $\sim$49--55\% vs.\ $\sim$38--49\%), indicating that a stronger questioner does not merely find more failures but raises the difficulty floor even on the hardest skills.

\begin{table}[h!]
\centering
\caption{Failure landscape for the 7B questioner across answerer scales. \textbf{Left:} Universally hard skills ranked by minimum FDR across all answerers. \textbf{Right:} Model-specific skills ranked by FDR standard deviation. FDR values are percentages.}
\label{tab:landscape_7b}
\resizebox{\textwidth}{!}{%
\begin{tabular}{lcccc||lccccc}
\toprule
\multicolumn{4}{c}{\textbf{Universally Hard Skills}} & & \multicolumn{5}{c}{\textbf{Model-Specific Skills}} \\
\cmidrule(lr){1-4} \cmidrule(lr){6-10}
\textbf{Skill} & \textbf{3B} & \textbf{7B} & \textbf{8B} & & \textbf{Skill} & \textbf{3B} & \textbf{7B} & \textbf{8B} & $\boldsymbol{\sigma}$ \\
\midrule
Logo Recognition              & 59.0 & 59.5 & 55.1 & & Intersection Detection        & 67.4 & 55.4 & 16.3 & 26.7 \\
Connector Connectivity        & 65.1 & 61.1 & 52.8 & & Referent Resolution           & 56.5 & 59.3 & 17.2 & 23.5 \\
Aerodynamic Feature Recog.    & 62.1 & 52.2 & 54.8 & & Geometric Measurement         & 57.5 & 61.8 & 22.6 & 21.5 \\
Traffic Sign Recognition      & 68.8 & 51.1 & 59.0 & & Image Center Localization     & 57.9 & 45.0 & 18.2 & 20.2 \\
Label Detection               & 61.3 & 54.1 & 50.4 & & Set Membership                & 64.8 & 39.4 & 25.7 & 19.8 \\
Attribute-Based Filtering     & 61.5 & 50.7 & 50.1 & & Adjacency Relations           & 50.5 & 50.0 & 16.3 & 19.6 \\
UI Recognition                & 54.9 & 52.4 & 50.0 & & Waveform Recognition          & 58.3 & 30.4 & 20.7 & 19.5 \\
Sign Recognition              & 60.2 & 49.6 & 60.6 & & Layout Recognition            & 61.8 & 61.1 & 28.4 & 19.1 \\
Table Row Extraction          & 75.0 & 48.5 & 52.2 & & 3D Shape Alignment            & 58.3 & 65.1 & 29.6 & 18.9 \\
Cross-Modal Grounding         & 51.6 & 51.4 & 48.4 & & Paint Identification          & 61.6 & 45.9 & 27.6 & 17.0 \\
\bottomrule
\end{tabular}%
}
\end{table}

\noindent \textbf{Model-Specific Skills.} In contrast, several skills exhibit high variance in failure rates across answerer scales. Skills such as \textit{Fine-Grained Difference Detection}, \textit{Adjacency Relations}, and \textit{Image Completion} are highly challenging for the 3B and 7B answerers but are substantially mitigated in the 8B model. For instance, with the 3B questioner, \textit{Fine-Grained Difference Detection} drops from 58.2\% FDR on the 3B answerer to 12.2\% on the 8B answerer ($\sigma = 23.6$). Comparing the two tables also reveals an interesting interaction between questioner and answerer capacity: \textit{UI Recognition} appears as a model-specific skill under the 3B questioner (8B answerer FDR of 23.5\%), yet the 7B questioner is able to craft questions about the same skill that remain universally hard across all answerers (8B answerer FDR of 50.0\%). This shows that the questioner's capacity affects not only how many failures it finds, but also which skills it can test in a way that challenges even stronger models.

Together, these findings show that our framework can surface both universal failures that persist regardless of model capacity and model-specific failures. The interaction between questioner and answerer capacity further suggests that stronger questioners do not just find more failures, but can probe the same skills more effectively.

\section{Conclusion}

We introduced an RL-based framework for automatically discovering failure modes in VLMs. Unlike previous approaches that require significant manual intervention, ours trains a questioner agent fully automatically using a VLM-based verifier. Our experiments demonstrate that the RL-trained questioner substantially outperforms competitive baselines across various metrics. In particular, our RL variant experienced $16\%$ more failures than the best baseline method ($126$ for RL vs $109$ for ZeroShot). In addition, the discovered failures had a higher failure rate and were more diverse than the baselines. The RL-variants discovered $11$ exclusive skills not explored by any other baseline methods. Our analysis further revealed that the RL-based approaches generate increasingly complex questions over time, starting with simpler ones. We also observed that some of the newer skills are discovered later in the training, after exploiting the model's basic vulnerabilities. These findings generalize across model scales and architectures. Unlike static benchmarks, our framework dynamically targets the specific weaknesses of a given VLM, offering a scalable alternative for discovering model-specific failure modes.


\bibliographystyle{plainnat}
\bibliography{main_arxiv}






\end{document}